# Biomarker Clustering of Colorectal Cancer Data to Complement Clinical Classification

Chris Roadknight[a], Uwe Aickelin[a], Alex Ladas[a], Daniele Soria[a], J Scholefield[b] and Lindy Durrant[b]
a. School of Computer Science
b. Faculty of Medicine & Health Sciences
The University of Nottingham
Contact email: cmr@cs.nott.ac.uk

*Abstract*— In this paper, we describe a dataset relating to cellular and physical conditions of patients who are operated upon to remove colorectal tumours. This data provides a unique insight into immunological status at the point of tumour removal, tumour classification and post-operative survival. Attempts are made to cluster this dataset and important subsets of it in an effort to characterize the data and validate existing standards for tumour classification. It is apparent from optimal clustering that existing tumour classification is largely unrelated to immunological factors within a patient and that there may be scope for re-evaluating treatment options and survival estimates based on a combination of tumour physiology and patient histochemistry.

## I. Introduction

Colorectal cancer is the third most commonly diagnosed cancer in the world. Colorectal cancers start in the lining of the bowel and grow into the muscle layers underneath then through the bowel wall [3]. TNM staging involves the Classification of Malignant Tumours

- Tumour (T). Size of the tumor and whether it has invaded nearby tissue
- Nodes (N). The extent to which regional lymph nodes involved
- Metastasis (M). This is the spread of a disease from one organ or part to another non-adjacent organ.

4 TNM stages (I,II,III,IV) are generated by combining these three indicator levels and are allied with increasing severity and decreasing survival rates.

Treatment options include minor/major surgery, chemotherapy, radiotherapy but the correct treatment is heavily dependent on the unique features of the tumour which are summarised by the TNM staging. Choosing the correct treatment at this stage is crucial to both the patients survival and quality of life. A major goal of this research is to automatically optimize the treatment plan based on the existing data.

The data for this research was gathered by scientists and clinicians at City Hospital, Nottingham. The dataset we use here is made up of the 84 attributes for 462 patients. The attributes are generated by recording metrics at the time of tumour removal, these include:

- Physical data (age, sex etc)
- Immunological data (levels of various T Cell subsets)
- Biochemical data (levels of certain proteins)
- Retrospective data (post-operative survival statistics)
- Clinical data (Tumour location, size etc).

In the research into the relationship between immune response and tumour staging there has been some support of the hypothesis that the adaptive immune response influences the behavior of human tumors. In situ analysis of tumor-infiltrating immune cells may therefore be a valuable prognostic tool in the treatment of colorectal cancer [7]. The immune and inflammation responses appear to have a role to play in the responses of patients to cancer [8] but the precise nature of this is still unclear.

The goal of this research is to assess the clustering behavior of this dataset to see how best the data can be grouped, how many clusters it should be grouped into and what these clusters represent. It is hoped that this may eventually help clinicians to decide how best to treat patients based on their biological, chemical and physical attributes at the time of tumor resection . Clustering is the act of assigning a set of n- dimensional attribute sets to clusters so that the members of one cluster are more similar to each other than to those in other clusters. If we define how patients can be optimally clustered into groups and how many groups they should clustered into, we can then assess how well these clusters match to key metrics such as survival and physical tumour grading.

This paper solely discusses unsupervised methods to represent the dataset. By removing supervised guidance we are allowing for undirected analysis of the data to inform us of features that may otherwise be missed. We have made a parallel effort to perform supervised modeling but this beyond the scope of this paper. Supervised efforts on cancer biomarker datasets have previously shown interesting relationships in the data and a phenomena called "Anti-learning" where testable feature detection using standard learning methods appears impossible when modeling of low sample sizes in high dimensional feature spaces is attempted [1,2].

## II. DATA ANALYSIS

The dataset supplied is a biological dataset and as such has a rich complement of preprocessing issues. 11.32% of the values are missing, with some attributes having over 40% missing values and some patients having over 30% missing values. Saying this, it is still an invaluable record of an extensive set of attributes from a relatively large number of patients.

Missing data poses a problem for most modelling techniques. One approach would be to remove every patient or every attribute with any missing data. This would remove a large number of entries, some of which only have a few missing values that are possibly insignificant. Another approach is to average the existing values for each attribute and to insert an average into the missing value space. The appropriate average may be the mean, median or mode depending on the profile of the data.

Much of the data takes the form of human analysis of biopsy samples stained for various markers. Rather than raw cell counts or measurements of protein levels we are presented with thresholded values. For instance, CD16 is found on the surface of different types of cells such as natural killer, neutrophils, monocytes and macrophages. The data contains a simple 0 or 1 for this rather than a count of the number of cells. This kind of manual inspection and simplification is true for most of the data and any modeling solution must work with this limitation.

It is apparent that there are some existing strong correlations in the data. By using a combination of correlation coefficients and expert knowledge the data was reduced down to a set of ~50 attributes. This included removing several measurements that were hindsight dependent (ie. chemo or radio treatment) and correlated with TNM stage. (ie. Dukes stage).

Single attribute relationships exist within the dataset but are not strong. Analysis of single attributes can yield a greater than 65% prediction rate when attempting to predict which TNM stage a patient was classified as but only ~55% when the TNM stages were restricted to the more interesting (TNM stage 2 or 3). If we look at CD59a and CD46 threshold values we can see that they are loosely related to survival (figure 1) with elevated levels of each indicating a reduction in survival averaging ~13 and 6 months. (Fig. 1a and b)

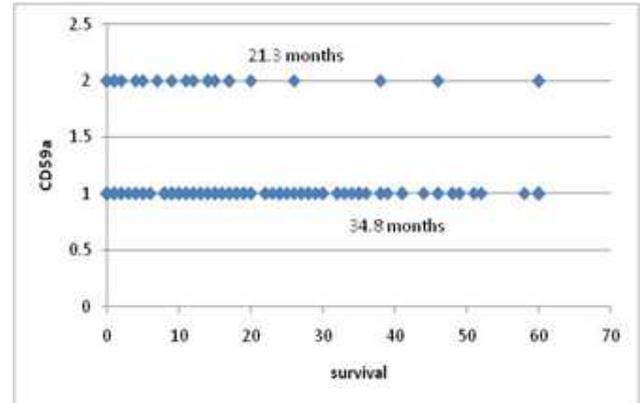

Figure 1a. Relationship of CD59 to survival with average survival rates.

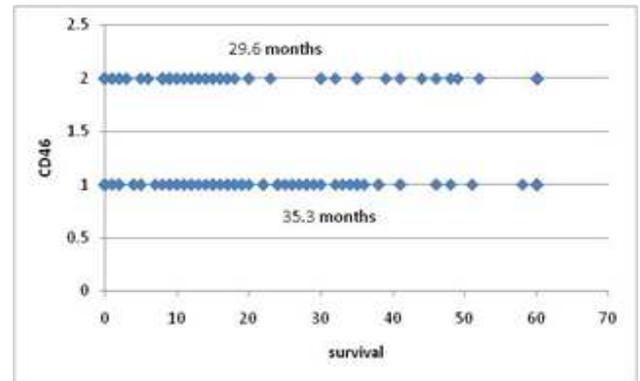

Figure 1b. Relationship of CD46 to survival with average survival rates.

## III. UNSUPERVISED LEARNING AND OPTIMAL CLUSTER NUMBERS

We initially look at how TNM stages are represented in the whole dataset when it is optimally clustered using a k-means approach [4]. This aims to divide multidimensional data into k clusters where each data point belongs to the cluster with the nearest mean. We quickly see that clustering based on biochemistry and physical attributes does not classify patients into the same classes as the TNM stages. Table 1 is a truth table for TNM stages 2 and 3 and a 2 cluster k-means approach showing that each cluster has a very similar number of patients that were TNM stage 2 and 3.

Table 1. Relationship of patients of 2 TNM categories to a 2 cluster k-means approach

|         | TNM Catagory | |
|---------|--------|--------|
| Cluster | 2 | 3 |
| 1 | 24.62% | 23.10% |
| 2 | 28.27% | 24.01% |

If we look at the survival rates for patients in each of the 2 clusters we see there is very little difference at averages of 41.75 and 41.98 months respectively, as opposed to 46.78 and 36.36 months for TNM stage 2 and 3 patients respectively. This would suggest that when we optimally cluster the biochemical data into 2 clusters, membership of the resulting clusters is a poor indicator of survival.

If we look at optimal 3 and 4 cluster solutions we get a much wider difference in survival periods for each cluster. Average survival periods for the three cluster solution were 39.56, 40.89 and 44.29 months and for the 4 cluster solution were 36.87, 40.15, 41.43 and 46.68 months. The TNM stages in each cluster are evenly spread suggesting survival can be predicted as well based on the biochemical marker combinations as the TNM stages. Figure 2a and 2b shows 2 of the 4 clusters, cluster 2 has the lowest survival rate at 36.87 months and cluster 4 has the highest survival rates at 46.68 months. If we look at the most widely differing attributes we can see that there are strong differences in the representation of these attributes in each cluster when compared to the average. To put this into context, nearly 60% of patients in cluster 2 had an elevated level for FLIP while only 21% of patients in cluster 4 had elevated levels of FLIP. FLIP is a key regulator of colorectal cancer cell death [10] so this wouldn't be surprising as the presence of it in patients may indicate more aggressive tumours.

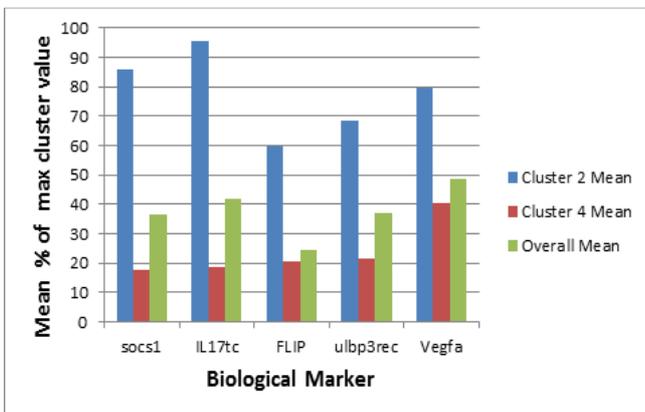

Figure 2a. Mean values for key markers in cluster 2 (low survival rates) and cluster 4 (highest survival rates)

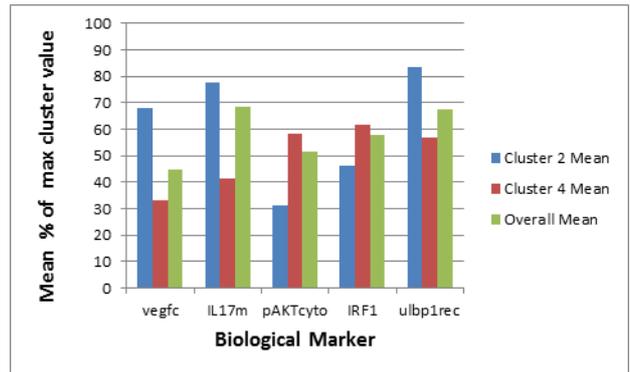

Figure 2b. Mean values for key markers in cluster 2 (low survival rates) and cluster 4 (highest survival rates)

IV. CLUSTER ANALYSIS OF DATASETS

Using research by Soria *et al* [5] as a basis we attempted to discover the optimal number of classes patients with different TNM stage tumours should be classified into. We had been largely unsuccessful in classifying patients into their TNM stages based on their attribute measurements and one possible reason for this is that the patients actually belong to 3 or more classes and attempting to classify them in a binary way resulted in poor performance.

In particular we implement a shorter version of the algorithmic framework, proposed by Soria et al [9], which combines different clustering algorithms and, with the use of Validation Indices, tries to define the optimal number of classes that describe the dataset. After this a method called Consensus Clustering is used to define the classes. The proposed framework consists of up to 7 steps, of which we will use the first 4, these are:

1. Preprocessing : This step includes the deletion of rows which contain missing values, data transformation and the calculation of descriptive statistics.

2. Clustering : In this step various clustering algorithms are applied to the dataset including Hierarchical Clustering, K-means, PAM and Fuzzy C-means.

3. Validation : In this step the utilisation of validation indices helps us to find the optimal number of clusters when this number is not known before the analysis.

4. Visualization & Agreement : In this step we obtain a general characterisation of the cluster analysis we performed here through various plots

We began by taking a small subset of the data that had very few missing values (~0.5%) and only pertained to TNM stage 2 and 3 patients, replacing any missing values with the modal value. The main aim here was to define how many clusters would be required to classify the data optimally and if the resulting clusters resembled the TNM staging. A k-means clustering approach was then used on the dataset to optimally cluster the data into 2-15 clusters.

Each of these clustering approaches was then tested to see how well the clusters fitted using 6 clustering indices to score them [6]. Results shown in figure 3 show that 3 or 4 clusters appears to be optimal for all 6 Indexes

We then ranked the performance of the six cluster indexing approaches for 2 to 10 clusters and calculated a value to represent performance using the reciprocal of the average ranking performance, with and without the Friedman Index (this index appeared to be the least reliable). The results from this (figure 4) would suggest that patients with TNM stage 2 and 3 tumours should be separated into 3 or 4 groups based on their immunology and chemistry statistics.

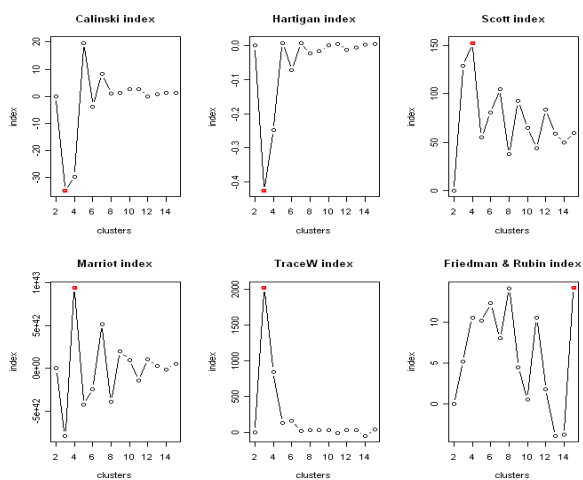

Figure 3. Cluster Optimality as defined by 6 cluster index calculations

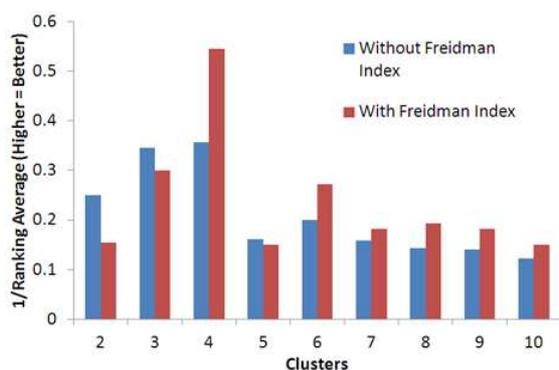

Figure 4. Ranking calculation for 2 – 10 clusters

If we look in more detail at the cluster membership we can get an idea about how attributes are divided up into 3clusters (table 2). Discussing the role of each abbreviated attribute presented here is beyond the scope of this paper but it includes levels for various immune cell subsets (eg. CD68), tumour suppression proteins (eg. P53) and degree of mutation in sections of DNA (MSI). These results continue to confirm that the patients are optimally separated into greater than 2 groups based on their immunohistochemistry.

Table 2. Levels of cell markers for each cluster as a percentage above (+) or below (-) average

|            | Cluster 1 | Cluster 2 | Cluster 3 |
|------------|-----------|-----------|-----------|
| CD68       | -38.03    | -9.09     | 44.59     |
| vegfc      | -26.89    | -9.28     | 34.23     |
| micahilo   | -25.37    | 17.39     | 7.55      |
| apcMHC2    | -4.81     | -21.16    | 24.57     |
| p53        | 17.42     | -24.09    | 6.32      |
| ulbp3rec   | -15.04    | -10.01    | 23.71     |
| ulbp1rec   | -13.07    | -11.81    | 23.55     |
| CD16       | -5.43     | -18.01    | 22.19     |
| bcl2       | 1.41      | -17.46    | 15.20     |
| DR4        | -1.31     | 16.29     | -14.19    |
| msi        | -15.01    | 2.60      | 11.74     |
| p27nuc     | -13.16    | 4.45      | 8.25      |
| ulbp2rec   | -10.63    | -3.08     | 12.96     |
| mhc        | -11.59    | 2.25      | 8.83      |
| FLIP       | -11.41    | 2.43      | 8.50      |
| cd46       | 8.47      | -4.11     | -4.13     |
| p27cyto    | -6.42     | -1.39     | 7.39      |
| stat1nuc2  | 0.62      | -4.41     | 3.59      |
| ki67       | -1.61     | -2.87     | 4.23      |
| cd59a      | -4.05     | 0.99      | 2.89      |
| mhcallele  | 0.92      | 2.17      | -2.92     |

Next we look at the relationship between the inclusion of missing values with predicted number of clusters. We have reported in detail the findings from two processed datasets that point to the data being optimally clustered into 3 datasets. In this section we analyse this finding further to see how the optimal number of clusters changes with dataset size and if there is any effect of adding data from patients with TNM stage 1 and 4 tumours

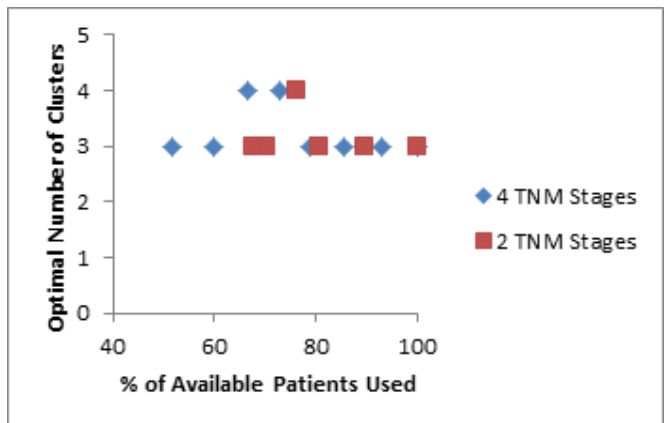

Figure 5. Optimal cluster number using different patient subsets.

If we reduce the number of patients based on removing those with the most missing data points first, and then look at the optimal number of clusters for each resulting dataset we see a graph that suggests 3 or 4 clusters solutions are optimal for datasets with just TNM stage 2 and 3 tumours (figure 5).. This would suggest that the TNM staging may not a powerful reflection of physical and biochemical conditions of a patient at the time of operation.

We can also look at the effect of removing attributes from the dataset one at a time based on missing values and seeing how this affects the optimal number of clusters for both the 2 and 4 TNM scenarios (Figures 6a and 6b). These show that 3 clusters to be most optimal, especially in the range of 25-35 attributes. This may be the most important range given that higher than this many more attributes with a high percentage of missing values are included and lower than 25 starts to exclude more and more important attributes. Overall it is shown that a dataset classified into 2 groups based on physical characteristics of the tumour does not agree with classification based on immune-histochemical data.

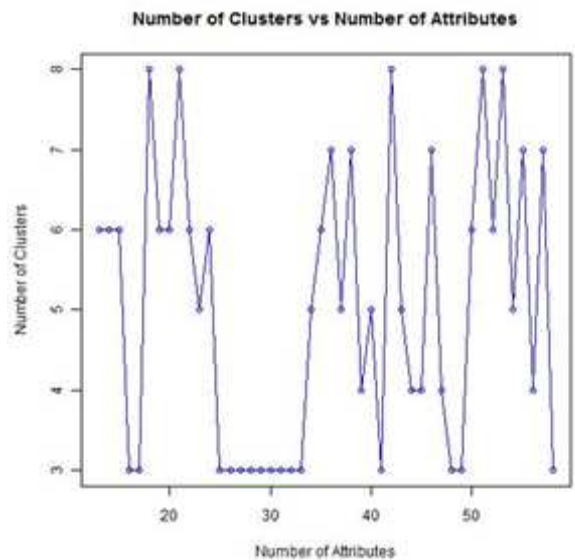

Figure 6a. Optimal number of clusters using consensus clustering for TNM 2,3

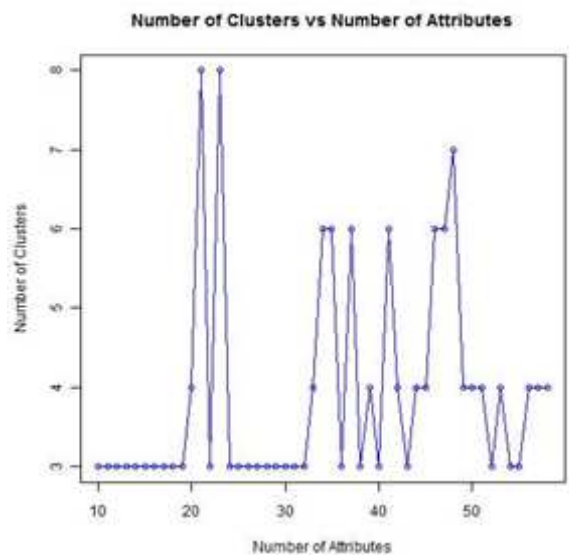

Figure 6b. TNM 1,2,3,4 tumour patients with an increasing number of attributes

## V. CONCLUSIONS

We have presented results for a unique dataset based on the biochemical and factors associated with colorectal tumour patients. This dataset is limited in many ways, but extremely important nonetheless and understanding any relationships or features based on the dataset to hand is an urgent priority. Currently, the most important guide to post-operative treatment is the TNM grading of the removed tumour. We argue here that TNM grading may not be aligned to the biochemical status of the patient. Given that the ongoing treatment for a patient immediately after tumor resection is directly acting upon a patients biochemistry it could be argued that an understanding of this immunological, biological and chemical status of the patient is at least as important as the physical characteristics of the tumour.

It is apparent that when clustering into 2 groups based on a selection of biochemical and physical characteristics is a poor representation of the patient state. There is no relationship between the new clusters and survival or TNM stage. This could be argued another way that 2 TNM stages for these patients is a poor representation of their biochemical status. When the number of clusters is examined it seems apparent that 3 or 4 clusters is optimal for patients, regardless of if they are from 4 TNM stages or just the most ambiguous TNM stage 2 and 3 group. Again, this would suggest that TNM staging is a poor representation of patients biochemical state.

The fact that prediction of survival can be achieved at comparable rates to TNM staging for 2 of the groups (36.67|36.87 and 46.68|46.78) suggests that survival is actually based on BOTH the physical metrics used for TNM staging AND the biochemical and immunological markers presented at the time of tumor resection.